\documentclass[conference]{IEEEtran}
\IEEEoverridecommandlockouts
\usepackage{cite}
\usepackage{amsmath,amssymb,amsfonts}
\usepackage{algorithmic}
\usepackage{graphicx}
\usepackage{textcomp}
\usepackage[table,x11names]{xcolor}
\usepackage{adjustbox}
\usepackage{booktabs}
\usepackage{multirow}
\usepackage{pifont}
\usepackage[ruled, vlined, noend]{algorithm2e}
\usepackage{stackengine}
\usepackage{makecell}
\usepackage{hyperref}

\def\BibTeX{{\rm B\kern-.05em{\sc i\kern-.025em b}\kern-.08em
    T\kern-.1667em\lower.7ex\hbox{E}\kern-.125emX}}

\DeclareMathAlphabet{\mathmybb}{U}{bbold}{m}{n}

\DeclareMathOperator*{\argmin}{arg\,min}

\newcommand{\cmark}{\ding{51}}%
\newcommand{\xmark}{\ding{55}}%

\definecolor{darkpastelgreen}{rgb}{0.01, 0.75, 0.24} %
\definecolor{darkpowderblue}{rgb}{0.0, 0.2, 0.6} %
\definecolor{darkpastelred}{rgb}{0.76, 0.23, 0.13} %
\definecolor{darkpastelblue}{rgb}{0.47, 0.62, 0.8} %
\definecolor{cinnamon}{rgb}{0.82, 0.41, 0.12} %
\definecolor{cinereous}{rgb}{0.6, 0.51, 0.48} %

\begin{document}

\title{FedDrive v2: an Analysis of the Impact of Label Skewness in Federated Semantic Segmentation for Autonomous Driving} 

\author{\IEEEauthorblockN{1\textsuperscript{st} Eros Fanì}
\IEEEauthorblockA{\textit{DAUIN Politecnico di Torino} \\
Turin, Italy \\
eros.fani@polito.it}
\and
\IEEEauthorblockN{2\textsuperscript{nd} Marco Ciccone}
\IEEEauthorblockA{\textit{DAUIN Politecnico di Torino} \\
Turin, Italy \\
marco.ciccone@polito.it}
\and
\IEEEauthorblockN{3\textsuperscript{rd} Barbara Caputo}
\IEEEauthorblockA{\textit{DAUIN Politecnico di Torino} \\
Turin, Italy \\
barbara.caputo@polito.it}
}

\maketitle

\begin{abstract}
We propose FedDrive v2, an extension of the Federated Learning benchmark for Semantic Segmentation in Autonomous Driving. While the first version aims at studying the effect of domain shift of the visual features across clients, in this work, we focus on the distribution skewness of the labels. We propose six new federated scenarios to investigate how label skewness affects the performance of segmentation models and compare it with the effect of domain shift. Finally, we study the impact of using the domain information during testing. 

\noindent Official website: \url{https://feddrive.github.io}

\end{abstract}


\vspace{-.3em}
\section{Introduction and Related Works}
An essential challenge for robust decision-making in autonomous vehicles such as self-driving cars is to design systems that can effectively gather and comprehend complex visual cues from their surroundings. A crucial vision task for perception is Semantic Segmentation (SS)~\cite {guo2018review}, which to be effectively trained and deployed requires collecting and annotating large datasets that cover the entire distribution of possible visual events. 
However, data collected by autonomous vehicles are generally covered by privacy regulations and cannot be freely shared with a central institution to train a centralized model. To tackle this issue, privacy-preserving approaches such as Federated Learning (FL)~\cite{fedavg} have been proposed as a possible solution to collaboratively train models across devices or data sources (clients) in a distributed fashion, while maintaining data confinement.

Despite the effectiveness of FL, a significant challenge in training through distributed learning arises from the \textit{statistical heterogeneity} among clients, leading to slower convergence of the global model. This is generally caused by the \textit{domain shift} of features within the same categories across clients or \textit{label distribution skewness}, which refers to an imbalanced or uneven distribution of labeled data across participating clients. Most methods primarily focus on addressing the label distribution skewness \cite{zhang2022federated, 9835537}. In contrast, only a few consider domain shift as a primary source of statistical heterogeneity~\cite{SiloBN, fedbn}. Additionally, these issues have been predominantly studied from a theoretical perspective, with limited emphasis on more structured tasks such as SS, addressing data heterogeneity among clients mainly for the classification task \cite{mime, scaffold, feddyn}, especially on small datasets. Only a handful of studies have ventured into large-scale visual classification~\cite{hsu2020federated}. 
Still, the research community has shown a growing interest in FL methods for autonomous driving~\cite{li2021privacy, du2020federated, jallepalli2021federated, tian2022federated}. However, the SS task has largely been overlooked in this context, with only a few notable exceptions \cite{feddrive, yao2022federated, ladd}. 

In particular, FedDrive \cite{feddrive} was the first benchmark of SS within the context of FL for autonomous vehicles, evaluating model generalization across various real-world conditions on synthetic and real-world federated datasets. While FedDrive mainly focuses on the domain shift problem, another critical source of statistical heterogeneity is label skewness. This arises because certain clients may have access to environments with different sets of categories and observe some of them more frequently than others. Indeed, a fleet of autonomous vehicles deployed in diverse locations may have access to a restricted or imbalanced set of classes. For instance, they might record varying numbers of riders, vehicles, traffic signs, or pedestrians, or might not encounter certain classes at all.

To study this problem, we introduce FedDrive v2, an expansion to the existing FedDrive benchmark \cite{feddrive} to evaluate the impact of label distribution skewness in distributed training. %
With FedDrive v2, we double the federated datasets introducing a new imbalanced federated split and novel training/test partition intended to investigate the domain shift challenge further. In addition, we analyze how style transfer and domain generalization techniques are affected by the label skewness problem. %
Finally, at inference time, we explore the impact of using local statistics for SiloBN \cite{SiloBN} on client-local test sets, resulting in up to 12 mIoU percentage points improvement. %

\vspace{-.3em}
\section{Federated Datasets description}

FedDrive v2 introduces a new train/test partition and a novel client distribution, generating six new federated datasets, thus doubling the scenarios already present in FedDrive \cite{feddrive}.

We build on the same federated setting and \textit{centralized datasets} of FedDrive: Cityscapes \cite{Cityscapes}, and IDDA \cite{idda}. Cityscapes is naturally divided into 2975 annotated images for training and 500 for testing, all gathered from similar cities from Central Europe in optimal weather conditions, while IDDA is a synthetic dataset with 105 different \textit{domains} of 60 images each, uniquely characterized by the triad (\textit{weather}, \textit{viewpoint}, \textit{town}), where \textit{weather}, \textit{viewpoint} and \textit{town} are one among three weather conditions, five different points of view for the camera recording the scenes simulating various vehicles, %
and six cities and one bucolic country, respectively.

\textbf{New Bus training/test partition.} For IDDA, the original benchmark already provided two different training/test partitions, both made of one training set and two test sets, the \textit{seen} and \textit{unseen}. Specifically, FedDrive introduced the \textit{Country} and the \textit{Rainy} partitions, where the unseen test set consists of all the images from the bucolic country domains for the former and rainy domains for the latter. The rainy weather condition and the bucolic country choices were made to exacerbate the semantic and appearance shift between the training and test images. However, the viewpoint axis, which may constitute another possible source of mismatch between training and test images, has been ignored. Here, we introduce a new unseen test set, \textit{Bus}, where the unseen test set has all the images from the bus domains. We chose the viewpoint of the bus since it provides a point of view set from above that is sensibly different from one of the other available vehicles, to maximize the discrepancy between the training and test images. In all these three partitions, the seen test set has 12 randomly sampled images from each domain left outside the unseen test set. Having two test sets of this nature is a simple yet effective choice to estimate the generalization capabilities of the methods on both seen and unseen environmental conditions. The remaining images constitute the training set and are divided across clients following different distributions (as explained below) to create the FL environment. 

\textbf{New Class Imbalance client distribution.} FedDrive proposes two client distributions, the \textit{Uniform} and \textit{Heterogeneous}, to study the effects of the statistical heterogeneity caused by domain shift. In the Uniform distribution, each client possesses random images for Cityscapes or one random image from each domain of the training set for IDDA. On the contrary, in the Heterogeneous, each client has access to images from a single city (for Cityscapes) or domain (for IDDA). The heterogeneity of this distribution mainly derives from the domain shift, based on the agreeable assumption that different cities or domains have various visual features. Potentially, it derives from the \textit{quantity skewness} too for the Cityscapes setting only, where the dimensions of the datasets of the clients are diverse. However, another realistic source of statistical heterogeneity not deemed by FedDrive is the label skewness since some clients may access environments where specific categories are more frequent than others or some clients are prevented from accessing these categories at all. \
Therefore, we introduce a novel client distribution, the \textit{Class Imbalance}, to analyze the behavior of the methods already studied in FedDrive in the presence of label skewness. 
We generate the distribution by proposing a general algorithm that maximizes the label skewness across clients. This is done iteratively by allocating images with the least frequent classes to clients, continuously keeping track of the images where each class appears at least once, and then moving to the new least frequent class, until all the images have been allocated to the clients. Our algorithm allows generating clients of any dimension to simultaneously introduce quantity skewness. We formally describe the procedure in Algorithm I.

Updated statistics of the federated datasets, including the Class Imbalance client distribution and the \textit{Bus} setting, are provided in Table \ref{tab:dataset_stats}. In all tables, we mark with $(^*)$ records taken from \cite{feddrive}. Moreover, Figure 1 shows a comparison of the class distribution among the clients per each client split for the Rainy training/test partition for IDDA.

\begin{algorithm}[t]
\scriptsize
\SetAlgoLined

\NoCaptionOfAlgo
\caption{\scriptsize\textsc{Algorithm I: Class Imbalance client distribution generation}} 
\textbf{Initialize:} \\
$\mathcal{D} = \text{set of all the training images}$ \\
Ordered set of empty client datasets $C$ \\
Ordered set $S: |S| = |C|$, $\sum_{s\in S} s = |\mathcal{D}|$, $s_i = \text{desired \# of samples $(x, y)$ for client }i$ \\
$\mathcal{D}_c \subseteq \mathcal{D}$: set of all $(x, y) \in \mathcal{D}$ such that class $c$ appears in $y$ \\

\For{\textbf{each} $s_i \in S$}{
    $\mathcal{X} = \argmin_{\mathcal{D}_c} |\mathcal{D}_c|$ \\
    \While{$|C_i| < s_i$} {
        Extract a subset $\mathcal{E}$ of $min(s_i - |C_i|, |\mathcal{X}|)$ uniformly sampled image and ground truth pairs $(x, y)$ from $\mathcal{X}$  \\
        $C_i = C_i \cup \mathcal{E}$ \\
        $\mathcal{D}_c = \mathcal{D}_c \setminus \mathcal{E}$ $\forall c$ \\
    }
}
\textbf{return} $C$ \\
\label{alg:class_imb}
\end{algorithm}

\begin{figure*}
    \centering
    \includegraphics[width=.89\linewidth]{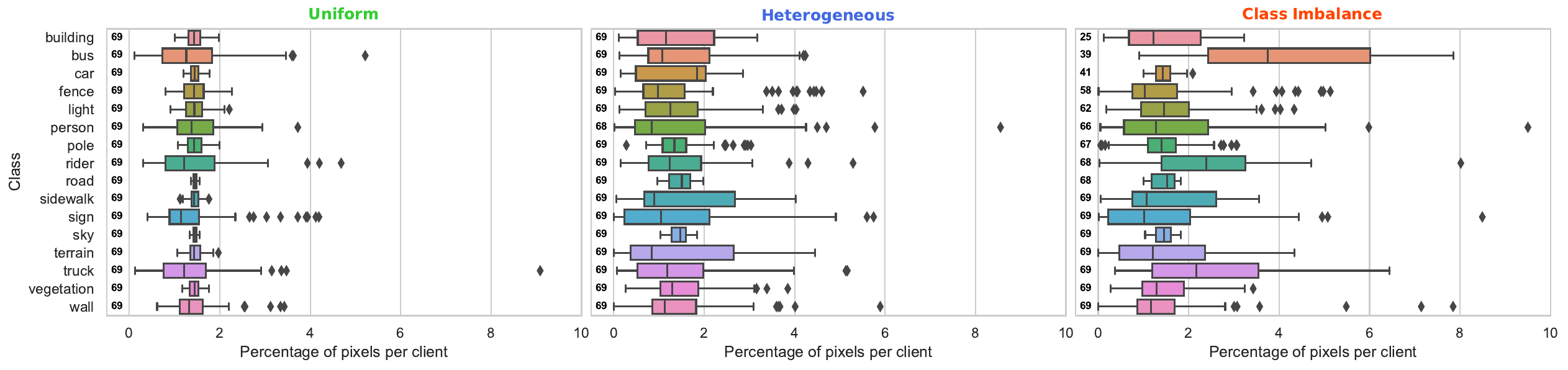}
    \label{fig:idda_class}
    \vspace{-1em}
    \caption{Class distribution comparison for the \color{cinereous} Rainy \color{black} IDDA training/test partition. For each class $c$, we evaluate the distribution of the number of $c$ pixels in each client that possesses at least one image with a pixel of class $c$, relative to the total number of pixels of class $c$ among all the clients. The numbers near the boxes are the number of clients having at least one image with that class, over 69 total clients. The Class Imbalance distribution has wider boxes, meaning there is more variation in the quantity of pixels of each class among the clients. \color{cinnamon} Country \color{black} and \color{darkpastelblue} Bus \color{black} provided similar cues on the class distributions.}
    \vspace{-2em}
\end{figure*}

\begin{table}
\vspace{-1.5em}
\caption{\scriptsize Summary of the FedDrive v2 Federatred Datasets.}
\centering
\vspace{-1em}
\label{tab:dataset_stats}
\begin{adjustbox}{width=\linewidth}

    \begin{tabular}{lccccc}

        \toprule
        
        \textbf{Dataset} & \textbf{Setting} & \textbf{Distribution} & \textbf{\# Clients}  & \textbf{\# img/cl}& \textbf{Test clients} \\
        
        \midrule
        
        \multirow{1}{*}{Cityscapes \cite{Cityscapes}} & - & \makecell{\color{darkpastelgreen} *Uniform, \color{darkpowderblue} *Heterogeneous, \\ \color{darkpastelred} Class Imbalance} & 146 & 10 - 45 &  unseen cities \\ 
        \midrule
        \multirow{3}{*}{IDDA \cite{idda} \vspace{-2.1em}} & \multirow{1}{*}{\color{cinnamon} Country} & \makecell{\color{darkpastelgreen} *Uniform, \color{darkpowderblue} *Heterogeneous, \\ \color{darkpastelred} Class Imbalance} & 90 & 48 & \makecell{seen + unseen \\ (country) domains} \\
         & \multirow{1}{*}{\color{cinereous}Rainy} &  \makecell{\color{darkpastelgreen} *Uniform, \color{darkpowderblue} *Heterogeneous, \\ \color{darkpastelred} Class Imbalance} & 69 & 48 & \makecell{seen + unseen \\ (rainy) domains} \\ 
         & \multirow{1}{*}{\color{darkpastelblue} Bus} &  \makecell{\color{darkpastelgreen} Uniform, \color{darkpowderblue} Heterogeneous, \\ \color{darkpastelred} Class Imbalance} & 83 & 48 & \makecell{seen + unseen \\ (bus) domains} \\
         
        \bottomrule
        
    \end{tabular}

\end{adjustbox}

\vspace{-1.5em}
\end{table}

\vspace{-.3em}
\section{Inference Strategies} \label{inference}
FedDrive examined the use of SiloBN \cite{SiloBN} as a method to counter domain shift in FL. They studied its effectiveness in the context of self-driving cars, specifically for the SS task. SiloBN enables the federated training of a model with Batch Normalization (BN) layers, even in cases of statistical heterogeneity, by keeping the BN statistics local to the clients. At inference time, the \textit{Standard strategy} is to compute the BN statistics directly from the test set. However, it is reasonable to assume that each training client has its local test set. We simulated this eventuality for the IDDA Heterogeneous federated datasets with the \textit{By Domain strategy}. We assign each image of the seen test set to the client of its corresponding domain. Then, the model is evaluated over all the local test sets by using the corresponding local statistics.

\vspace{-.3em}
\section{Experiments}
We run all the new experiments using a Tesla V100-SXM2-16GB. The chosen lightweight architecture is BiSeNet V2 \cite{bisenetv2}. The hyper-parameters choice is the same as the original FedDrive paper \cite{feddrive}. In the experiments with the server optimizer, we report only the results with the best server learning rate between 0.1 and 1.0.

\subsection{Cityscapes results}

Table \ref{tab:cts_classimb} shows the Cityscapes results with the Class Imbalance distribution. SiloBN still improves the performance, but the gains are less pronounced with respect to the Heterogeneous one since SiloBN mainly tackles the domain shift rather than the label skewness. Analogously, CFSI \cite{cfsi} and LAB \cite{lab} seem not to be particularly helpful when combined with SiloBN. However, with this client distribution, CFSI improves the performance by four percentage points with respect to FedAvg alone in the experiments without SiloBN.

Table \ref{tab:cts_opt} compares four server optimizers at different client distributions. Adam %
consistently performs better than the other server optimizers, and SGD is always the second best. Maybe surprisingly, FedAvgM struggles to achieve good performance.

\begin{table}
\centering
\makebox[0pt][c]{\parbox{\linewidth}{%
    \begin{minipage}{.5\linewidth}\centering
    
    \caption{\scriptsize Cityscapes results, \color{darkpastelred}Class Imbalance \color{black} split.}
    \label{tab:cts_classimb}
    \vspace{-1em}
    \begin{adjustbox}{width=.9\linewidth}
        \begin{tabular}{lccc}
            \toprule
            \textbf{Method} & \textbf{CFSI} & \textbf{LAB} & \textbf{mIoU $\pm$ std (\%)} \\
            \midrule
              \multirow{3}{*}{FedAvg} & \xmark & \xmark & 44.48 $\pm$ 1.70          \\
               & \cmark & \xmark & \textbf{48.28 $\pm$ 1.83}          \\ 
               & \xmark & \cmark & 44.34 $\pm$ 1.73          \\ 
               \cmidrule{1-4}
              \multirow{3}{*}{SiloBN} & \xmark & \xmark & \textbf{51.78 $\pm$ 1.15} \\
               & \cmark & \xmark & 50.82 $\pm$ 1.08          \\
               & \xmark & \cmark & 51.48 $\pm$ 1.22          \\
            \bottomrule
        \end{tabular}
    \end{adjustbox}
    
    \end{minipage}
    \hfill
    \begin{minipage}{.5\linewidth}\centering

    \caption{\scriptsize Cityscapes, server optimizers comparison.}
    \label{tab:cts_opt}
    \vspace{-1em}
    \begin{adjustbox}{width=.9\linewidth}
        \begin{tabular}{lcc}
            \toprule
            
            & \textbf{SGD} & \textbf{FedAvgM} \\
            
            \midrule
            
            \color{darkpastelgreen} Uniform & *45.62 $\pm$ 1.25 & 40.04 $\pm$ 4.26 \\
            \color{darkpowderblue} Heterogeneous & *43.33 $\pm$ 1.66 & 37.83 $\pm$ 4.61 \\
            \color{darkpastelred} Class Imbalance & 44.48 $\pm$ 1.70 & 36.13 $\pm$ 4.52 \\

            \midrule[.1em]

            & \textbf{Adam} & \textbf{AdaGrad} \\

            \midrule
            
            \color{darkpastelgreen} Uniform & \textbf{45.91 $\pm$ 1.28} & 44.30 $\pm$ 3.66 \\
            \color{darkpowderblue} Heterogeneous & \textbf{45.08 $\pm$ 1.55} & 42.28 $\pm$ 3.12 \\
            \color{darkpastelred} Class Imbalance & \textbf{45.21 $\pm$ 1.81} & 39.78 $\pm$ 4.10 \\
            
            \bottomrule
        
        \end{tabular}
    \end{adjustbox}
        
    \end{minipage}
    \hfill
}}

\vspace{-2em}

\end{table}

\subsection{IDDA results} \label{idda_results}

Table \ref{tab:idda_fedavg} shows the results for the FedAvg experiments on the IDDA dataset for every proposed scenario, eventually applying CFSI and LAB style translation techniques. First, we observe that style translation techniques are always helpful since they improve the performance or perform the same as FedAvg alone in the worst case, as for the Class Imbalance Rainy experiments. Moreover, we observe LAB outperforms CFSI in all the Class Imbalance experiments. On the contrary, this is not always true for the Heterogeneous experiments, where LAB is superior only for the Bus ones. Finally, the Class Imbalance experiments perform much better than the related Heterogeneous experiments, meaning that the maximum label skewness you can achieve in this SS task only slightly contributes to the statistical heterogeneity, and you can reach higher statistical heterogeneity from the domain shift.

\begin{table}[t]

    \caption{%
    \scriptsize IDDA, FedAvg experiments.}
    \vspace{-1em}
    \label{tab:idda_fedavg}
    \centering
    \begin{adjustbox}{width=.55\columnwidth}
    \centering
    
    \begin{tabular}{lccccc}
    
    \toprule
    
    & & \textbf{CFSI} & \textbf{LAB} & \textbf{Unseen} & \textbf{Seen} \\
    
    \midrule
    
    \rowcolor{gray!15}\multirow{3}{*}{\rotatebox[origin=c]{90}{\color{darkpastelgreen} Uniform}} & \multirow{1}{*}{\color{cinnamon} *Country} &  \xmark & \xmark & 49.74 $\pm$ 0.79 & 63.57 $\pm$ 0.60 \\
    \rowcolor{gray!15}& \multirow{1}{*}{\color{cinereous} *Rainy} & \xmark & \xmark & 27.61 $\pm$ 2.80 & 62.72 $\pm$ 3.65 \\
    \rowcolor{gray!15}\multirow{-3}{*}{\rotatebox[origin=c]{90}{\color{darkpastelgreen} Uniform}}& \multirow{1}{*}{\color{darkpastelblue} Bus} & \xmark & \xmark & 58.51 $\pm$ 1.32 & 64.87 $\pm$ 0.65 \\

    \midrule
    
    \multirow{9}{*}{\rotatebox[origin=c]{90}{\color{darkpowderblue} Heterogeneous\hspace{1.2em}}} & \multirow{3}{*}{\color{cinnamon} *Country} & \xmark & \xmark & 40.01 $\pm$ 1.26 & 42.43 $\pm$ 1.78 \\
    & & \cmark & \xmark & \textbf{45.70 $\pm$ 1.73} & \textbf{54.70 $\pm$ 1.12} \\
    & & \xmark & \cmark & 45.68 $\pm$ 1.04 & 56.59 $\pm$ 0.90 \\
    \cmidrule{2-6}
    & \multirow{3}{*}{\color{cinereous} *Rainy} & \xmark & \xmark & 26.75 $\pm$ 2.32 & 38.18 $\pm$ 1.40 \\
    & & \cmark & \xmark & \textbf{31.05 $\pm$ 2.68} & 55.24 $\pm$ 1.65 \\
    & & \xmark & \cmark & 26.82 $\pm$ 1.78 & \textbf{58.85 $\pm$ 0.89} \\
    \cmidrule{2-6}
    & \multirow{3}{*}{\color{darkpastelblue} Bus} & \xmark & \xmark & 38.13 $\pm$ 1.96 & 45.71 $\pm$ 1.65 \\
    & & \cmark & \xmark & 48.88 $\pm$ 1.46 & 56.93 $\pm$ 1.39 \\
    & & \xmark & \cmark & \textbf{50.48 $\pm$ 1.09} & \textbf{58.84 $\pm$ 0.97} \\

    \midrule
    
    \multirow{9}{*}{\rotatebox[origin=c]{90}{\color{darkpastelred} Class Imbalance\hspace{1.2em}}} & \multirow{3}{*}{\color{cinnamon} Country} & \xmark & \xmark & 47.58 $\pm$ 0.69 & 58.09 $\pm$ 0.78 \\
    & & \cmark & \xmark & 48.69 $\pm$ 0.82 & 59.67 $\pm$ 0.88 \\
    & & \xmark & \cmark & \textbf{48.91 $\pm$ 0.78} & \textbf{60.07 $\pm$ 1.18} \\
    \cmidrule{2-6}
    & \multirow{3}{*}{\color{cinereous} Rainy} & \xmark & \xmark & \textbf{29.46 $\pm$ 1.90} & 58.75 $\pm$ 1.45 \\
    & & \cmark & \xmark & 25.69 $\pm$ 2.77 & 60.37 $\pm$ 0.60 \\
    & & \xmark & \cmark & 29.11 $\pm$ 1.68 & \textbf{61.22 $\pm$ 0.98} \\
    \cmidrule{2-6}
    & \multirow{3}{*}{\color{darkpastelblue} Bus} & \xmark & \xmark & 53.29 $\pm$ 2.05 & 60.47 $\pm$ 1.75 \\
    & & \cmark & \xmark & 52.91 $\pm$ 1.33 & 61.24 $\pm$ 1.24 \\
    & & \xmark & \cmark & \textbf{54.01 $\pm$ 1.15} & \textbf{62.10 $\pm$ 0.55} \\
    
    \bottomrule
    
    \end{tabular}
    \end{adjustbox}
    \vspace{-5pt}
    
\end{table}

Table \ref{tab:idda_h_silobn} and Table \ref{tab:idda_ci_silobn} report the SiloBN experiments for the IDDA Heterogeneous and Class Imbalance federated datasets, respectively. LAB experiments consistently outperform most of the other experiments in most scenarios, also in this case. The By Domain inference strategy for the Heterogeneous experiments is especially good without style translation techniques. Additionally, if we compare these results with the ones in Table \ref{tab:idda_fedavg}, we can observe that SiloBN + LAB is often the best combination of techniques.

\begin{table}[t]
    \vspace{-1em}
    \caption{\scriptsize IDDA \color{darkpowderblue} Heterogeneous \color{black} results, SiloBN experiments.%
    }
    \label{tab:idda_h_silobn}
    \vspace{-1.0em}
    
    \centering
    \begin{adjustbox}{width=.65\columnwidth}
    \centering
    
    \begin{tabular}{lccccc}
    
    \toprule
    
    & \multirow{2}{*}{\textbf{CFSI}} & \multirow{2}{*}{\textbf{LAB}} & \multirow{2}{*}{\textbf{Unseen}} & \multicolumn{2}{c}{\textbf{Seen}} \\
    & & & & \textbf{Standard} & \textbf{By Domain} \\

    \midrule
    
    \multirow{3}{*}{\rotatebox[origin=c]{90}{\color{cinnamon} Country}} & \xmark & \xmark & *45.32 $\pm$ 0.90 & 54.46 $\pm$ 0.72 & *58.82 $\pm$ 2.93 \\
    & \cmark & \xmark & *49.17 $\pm$ 1.01 & 63.43 $\pm$ 0.58 & *61.22 $\pm$ 3.88 \\
    & \xmark & \cmark & *\textbf{50.43 $\pm$ 0.63} & \textbf{64.59 $\pm$ 0.45} & *64.32 $\pm$ 0.76 \\
    \midrule
    \multirow{3}{*}{\rotatebox[origin=c]{90}{\color{cinereous} Rainy}} & \xmark & \xmark & *50.03 $\pm$ 0.79 & 54.36 $\pm$ 0.83 & *62.48 $\pm$ 1.42 \\
    & \cmark & \xmark & *50.54 $\pm$ 0.88 & 64.85 $\pm$ 0.72 & *63.04 $\pm$ 0.31 \\
    & \xmark & \cmark & *\textbf{53.99 $\pm$ 0.79} & \textbf{65.90 $\pm$ 0.55} & *65.85 $\pm$ 0.91 \\
    \midrule
    \multirow{3}{*}{\rotatebox[origin=c]{90}{\color{darkpastelblue} Bus}} & \xmark & \xmark & 47.37 $\pm$ 0.80 & 57.84 $\pm$ 0.89 & 61.56 $\pm$ 1.39 \\
    & \cmark & \xmark & 55.84 $\pm$ 0.99 & 65.78 $\pm$ 0.81 & 64.03 $\pm$ 2.68 \\
    & \xmark & \cmark & \textbf{56.23 $\pm$ 0.64} & \textbf{66.98 $\pm$ 0.34} & 66.23 $\pm$ 0.83 \\

    \bottomrule
    
    \end{tabular}
    \end{adjustbox}

\vspace{-2em}

\end{table}

\begin{table}[t]

    \caption{\scriptsize IDDA \color{red} Class Imbalance \color{black} results, SiloBN experiments.}
    \label{tab:idda_ci_silobn}
    \vspace{-1em}

    \centering
    \begin{adjustbox}{width=.5\columnwidth}
    \centering
    
    \begin{tabular}{lcccc}
    
    \toprule
    
    & \textbf{CFSI} & \textbf{LAB} & \textbf{Unseen} & \textbf{Seen} \\
    
    \midrule
    
    \multirow{3}{*}{\rotatebox[origin=c]{90}{\color{cinnamon} Country}} & \xmark & \xmark & 51.19 $\pm$ 0.55 & 66.46 $\pm$ 0.35 \\
    & \cmark & \xmark & 51.73 $\pm$ 0.69 & 67.22 $\pm$ 0.41 \\
    & \xmark & \cmark & \textbf{53.10 $\pm$ 0.55} & \textbf{67.33 $\pm$ 0.36} \\
    \midrule
    \multirow{3}{*}{\rotatebox[origin=c]{90}{\color{cinereous} Rainy}} & \xmark & \xmark & \textbf{54.68 $\pm$ 0.56} & 66.41 $\pm$ 0.53 \\
    & \cmark & \xmark & 54.35 $\pm$ 1.03 & 67.07 $\pm$ 0.39 \\
    & \xmark & \cmark & 52.91 $\pm$ 0.84 & \textbf{67.63 $\pm$ 0.32} \\
    \midrule
    \multirow{3}{*}{\rotatebox[origin=c]{90}{\color{darkpastelblue} Bus}} & \xmark & \xmark & 57.63 $\pm$ 0.59 & 67.54 $\pm$ 0.41 \\
    & \cmark & \xmark & 57.68 $\pm$ 0.66 & \textbf{67.89 $\pm$ 0.59} \\
    & \xmark & \cmark & \textbf{58.02 $\pm$ 0.55} & 67.80 $\pm$ 0.43 \\
    
    \bottomrule
    
    \end{tabular}
    
    \end{adjustbox}
    \vspace{-1em}

\end{table}

Finally, we analyzed the behavior of the server optimizers for IDDA. However, in Tables \ref{tab:idda_opt} and \ref{tab:idda_opt_silobn}, we show only the comparison of SGD with FedAvgM for the IDDA dataset because, contrary to the Cityscapes experiments, Adam and AdaGrad failed to achieve good performance in most of the experiments. The only two exceptions were two experiments without SiloBN that showed good performance on the unseen test set using Adam (namely, Uniform Rainy: unseen = $31.72 \pm 1.74 \%$, seen = $65.39 \pm 0.52 \%$; Class Imbalance Rainy: unseen = $35.78 \pm 1.93 \%$ mIoU, seen = $57.50 \pm 1.17 \%$ mIoU). In all the other cases, FedAvgM is always the best optimizer. For the SiloBN experiments, the By Domain inference strategy is always the best, improving the performance up to 12 mIoU percentage points for the Rainy FedAvgM experiment.

\begin{table}[t]

    \caption{%
    \scriptsize SGD vs FedAvgM comparison on IDDA, FedAvg experiments. %
    }
    \label{tab:idda_opt}
    \vspace{-1em}
    
    \centering
    \begin{adjustbox}{width=.78\linewidth}
    \centering
    
    \begin{tabular}{lccccc}
    
    \toprule
    
     & & \multicolumn{2}{c}{\textbf{SGD}} & \multicolumn{2}{c}{\textbf{FedAvgM}} \\
     & & \textbf{Unseen} & \textbf{Seen} & \textbf{Unseen} & \textbf{Seen} \\
    
    \midrule

    \multirow{3}{*}{\rotatebox[origin=c]{90}{\color{darkpastelgreen} Uniform}} & \color{cinnamon} *Country & 49.74 $\pm$ 0.79 & 63.57 $\pm$ 0.60 & \textbf{55.47 $\pm$ 1.07} & \textbf{71.27 $\pm$ 0.85} \\
     & \color{cinereous} *Rainy & 27.61 $\pm$ 2.80 & 62.72 $\pm$ 3.65 & 29.83 $\pm$ 2.03 & \textbf{70.99 $\pm$ 0.71} \\
     & \color{darkpastelblue} Bus & 58.51 $\pm$ 1.32 & 64.87 $\pm$ 0.65 & \textbf{62.32 $\pm$ 1.25} & \textbf{71.76 $\pm$ 0.37} \\

    \midrule
    
    \multirow{3}{*}{\rotatebox[origin=c]{90}{\color{darkpowderblue} Heter.}} & \color{cinnamon} *Country & 40.01 $\pm$ 1.26 & 42.43 $\pm$ 1.78 & \textbf{42.42 $\pm$ 2.15} & \textbf{44.38 $\pm$ 1.98} \\
     & \color{cinereous} *Rainy & 26.75 $\pm$ 2.32 & 38.18 $\pm$ 1.40 & \textbf{31.91 $\pm$ 3.77} & \textbf{41.21 $\pm$ 1.98} \\
     & \color{darkpastelblue} Bus & 38.13 $\pm$ 1.96 & 45.71 $\pm$ 1.65 & \textbf{40.39 $\pm$ 1.56} & \textbf{48.92 $\pm$ 1.54} \\

    \midrule

    \multirow{3}{*}{\rotatebox[origin=c]{90}{\color{darkpastelred} Cl. Imb.}} & \color{cinnamon} Country & 47.58 $\pm$ 0.69 & 58.09 $\pm$ 0.78 & \textbf{50.22 $\pm$ 1.17} & \textbf{63.01 $\pm$ 1.20} \\
     & \color{cinereous} Rainy & 29.46 $\pm$ 1.90 & 58.75 $\pm$ 1.45 & 32.30 $\pm$ 1.93 & \textbf{63.96 $\pm$ 1.11} \\
     & \color{darkpastelblue} Bus & 53.29 $\pm$ 2.05 & 60.47 $\pm$ 1.75 & \textbf{54.71 $\pm$ 1.65} & \textbf{64.45 $\pm$ 1.00} \\

    \bottomrule
    
    \end{tabular}
    \end{adjustbox}
    \vspace{-1.5em}
    
\end{table}

\begin{table}[t]

    \caption{%
    \scriptsize IDDA, \color{darkpowderblue} Heterogeneous\color{black}, server optimizers comparison using SiloBN. %
    }
    \label{tab:idda_opt_silobn}
    \vspace{-1em}
    
    \centering
    \begin{adjustbox}{width=.7\columnwidth}
    \centering
    
    \begin{tabular}{ccccc}
    
    \toprule
     
      \multirow{2}{*}{\textbf{Partition}} & \multirow{2}{*}{\textbf{Optimizer}} & \multirow{2}{*}{\textbf{*Unseen}} & \multicolumn{2}{c}{\textbf{Seen}} \\
      & & & \textbf{Standard} & \textbf{*By Domain} \\
    
    \midrule

    \multirow{2}{*}{\color{cinnamon} Country} & SGD & 45.32 $\pm$ 0.90 & 54.46 $\pm$ 0.72 & 58.82 $\pm$ 2.93 \\
     & FedAvgM & \textbf{46.20 $\pm$ 1.20} & 54.56 $\pm$ 1.29 & \textbf{61.99 $\pm$ 1.51} \\
    
    \midrule
    
    \multirow{2}{*}{\color{cinereous} Rainy} & SGD & \textbf{50.03 $\pm$ 0.79} & 54.36 $\pm$ 0.83 & 62.48 $\pm$ 1.42 \\
     & FedAvgM & 48.49 $\pm$ 1.04 & 51.71 $\pm$ 0.73 & \textbf{63.69 $\pm$ 1.25} \\

    \midrule

    \multirow{2}{*}{\color{darkpastelblue} Bus} & SGD & \textbf{47.37 $\pm$ 0.80} & 57.84 $\pm$ 0.89 & 61.56 $\pm$ 1.39 \\
     & FedAvgM & 46.91 $\pm$ 1.04 & 55.98 $\pm$ 0.76 & \textbf{ 63.41 $\pm$ 1.43 } \\
    
    \bottomrule
    
    \end{tabular}
    \end{adjustbox}
    \vspace{-1.5em}
    
\end{table}

\vspace{-.3em}
\section{Conclusion}

In this work, we extend FedDrive \cite{feddrive} by introducing a novel distribution of the clients to study the effect of label skewness and a new training/test partition for the IDDA dataset. We do so, by proposing an algorithm for generating class imbalanced splits for federated datasets. We also study the effect of a new inference strategy for SiloBN in the presence of test sets local to the clients. All our studies provide many additional experiments compared to FedDrive. We found that SiloBN \cite{SiloBN}, CFSI \cite{cfsi}, and LAB \cite{lab} could still be helpful in the presence of label skewness despite these techniques being designed for domain adaptation and generalization. In addition, results indicate that the domain shift is more challenging than label skewness in SS for autonomous vehicles. 
Future efforts will focus on designing specific algorithms to address class imbalance and label skewness issues in federated semantic segmentation. Additionally, we aim to enhance FedDrive with large-scale real-world datasets that mirror the long-tail distribution found in autonomous driving scenarios.

\vspace{-.3em}
\section*{Acknowledgements}

This study was carried out within the FAIR - Future Artificial Intelligence Research and received funding from the European Union Next-GenerationEU (PIANO NAZIONALE DI RIPRESA E RESILIENZA (PNRR) – MISSIONE 4 COMPONENTE 2, INVESTIMENTO 1.3 – D.D. 1555 11/10/2022, PE00000013). This manuscript reflects only the authors’ views and opinions, neither the European Union nor the European Commission can be considered responsible for them.

\vspace{-.3em}
\bibliographystyle{IEEEtran.bst} 
\bibliography{IEEEabrv,bibfile}

\end{document}